\documentclass[10pt,twocolumn,letterpaper]{article}

\usepackage{cvpr}
\usepackage{times}
\usepackage{epsfig}
\usepackage{graphicx}
\usepackage{amsmath}
\usepackage{amssymb}
\usepackage{algorithm}
\usepackage{algorithmicx}
\usepackage{algpseudocode}
\usepackage{spverbatim}
\usepackage{etoolbox}
\makeatletter
\preto{\@verbatim}{\topsep=1pt \partopsep=1pt}
\makeatother

\usepackage{booktabs}
\usepackage[x11names,dvipsnames,table]{xcolor} %for use in color links
\usepackage{multirow}
\usepackage{array}
\newcolumntype{L}[1]{>{\raggedright\let\newline\\\arraybackslash\hspace{0pt}}m{#1}}
\newcolumntype{C}[1]{>{\centering\let\newline\\\arraybackslash\hspace{0pt}}m{#1}}
\newcolumntype{R}[1]{>{\raggedleft\let\newline\\\arraybackslash\hspace{0pt}}m{#1}}

\usepackage[pagebackref=true,breaklinks=true,letterpaper=true,colorlinks,bookmarks=false]{hyperref}

\cvprfinalcopy % *** Uncomment this line for the final submission

\def\cvprPaperID{567} % *** Enter the CVPR Paper ID here

\ifcvprfinal\fi
\begin{document}

%%%%%%%%% TITLE
\title{Deep Collaborative Learning for Visual Recognition}
\author{Yan Wang\textsuperscript{1}, Lingxi Xie\textsuperscript{2},
Ya Zhang\textsuperscript{1}, Wenjun Zhang\textsuperscript{1}, Alan Yuille\textsuperscript{2} \\
\textsuperscript{1}School of Electronic Info. and Electrical Engi., Shanghai Jiao Tong University, Shanghai, China \\
{\tt\small tiffany940107@gmail.com}\quad{\tt\small yazhang@sjtu.edu.cn}\quad{\tt\small zhangwenjun@sjtu.edu.cn} \\
\textsuperscript{2}Center for Imaging Science, The Johns Hopkins University, Baltimore, MD, USA \\
{\tt\small 198808xc@gmail.com}\quad{\tt\small alan.l.yuille@gmail.com}
}

\maketitle
%\thispagestyle{empty}

%%%%%%%%% ABSTRACT
\begin{abstract}
Deep neural networks are playing an important role in state-of-the-art visual recognition.
To represent high-level visual concepts, modern networks are equipped with large convolutional layers,
which use a large number of filters and contribute significantly to model complexity.
For example, more than half of the weights of {\bf AlexNet} are stored in the first fully-connected layer ($4\rm{,}096$ filters).

We formulate the function of a convolutional layer as learning a large visual vocabulary,
and propose an alternative way, namely {\bf Deep Collaborative Learning} (DCL), to reduce the computational complexity.
We replace a convolutional layer with a two-stage DCL module,
in which we first construct a couple of smaller convolutional layers individually,
and then fuse them at each spatial position to consider feature co-occurrence.
In mathematics, DCL can be explained as an efficient way of learning compositional visual concepts,
in which the vocabulary size increases exponentially while the model complexity only increases linearly.
We evaluate DCL on a wide range of visual recognition tasks,
including a series of multi-digit number classification datasets,
and some generic image classification datasets such as {\bf SVHN}, {\bf CIFAR} and {\bf ILSVRC2012}.
We apply DCL to several state-of-the-art network structures,
improving the recognition accuracy meanwhile reducing the number of parameters ($16.82\%$ fewer in {\bf AlexNet}).
\end{abstract}

%%%%%%%%% BODY TEXT
\section{Introduction}
\label{Introduction}

Image classification is a fundamental problem in computer vision.
With the availability of large-scale image datasets~\cite{Deng_2009_ImageNet}
and powerful computational resources such as modern GPUs,
it is possible to train a convolutional neural network (CNN)~\cite{Krizhevsky_2012_ImageNet}
which significantly outperforms the conventional models like the Bag-of-Visual-Words~\cite{Csurka_2004_Visual}.

Most CNN architectures contain convolutional (or fully-connected) layers with a large number of filters,
which are designed to capture the increasing number of mid-level or high-level visual concepts.
These layers contribute significantly to model complexity.
As an example, the first fully-connected layer of {\bf AlexNet}~\cite{Krizhevsky_2012_ImageNet} contains $4096$ filters,
requiring $37.75\mathrm{M}$ parameters (more than $60\%$ of the parameters used in the entire network).
We formulate the function of a convolutional layer as learning a large visual vocabulary,
in which each filter is used to detect a specific visual concept via template matching.
Note that some previous work trains a large visual vocabulary~\cite{Jegou_2011_Product} using the composition of several small ones.
This idea is successfully applied to approximate nearest neighbor search~\cite{Ge_2013_Optimized}\cite{Zhang_2014_Composite},
image classification~\cite{Sanchez_2013_Image} and retrieval~\cite{Jegou_2012_Aggregating}.
We borrow this idea to reduce the computational complexity of the convolutional layers.

Our algorithm is named {\bf Deep Collaborative Learning} (DCL).
It is a generalized module which applies to a wide range of network structures.
The idea is very simple: a large convolutional layer can be simulated with the combination of several small convolutional layers.
As illustrated in Figure~\ref{Fig:DeepCollaborativeLearning}, DCL is a two-stage module to replace a convolutional layer.
At the first stage, we individually construct several convolutional layers with the same spatial resolution.
These {\em branches} are fused at the second stage,
which involves linear weighting followed by element-wise operation at each spatial position.
In mathematics, DCL can be explained as an efficient way of constructing a compositional visual vocabulary,
in which we spend linear complexity to increase the vocabulary size exponentially.

We evaluate DCL on a wide range of visual recognition tasks.
First, we generate a series of multi-digit number classification datasets
by pasting random {\bf MNIST} digits into a fixed spatial layout (see Section~\ref{Experiments:Structured} for details).
We keep the number of training images unchanged, although the number of categories grows exponentially.
DCL works better than conventional models because of two factors.
First, DCL enjoys a lower risk of over-fitting, especially in the scenario that the amount of training data is limited.
Second, as shown in visualization, DCL uses different branches to learn complementary visual concepts,
so that they can be combined to represent the large but decomposable set of visual categories.
We also apply DCL to some state-of-the-art network structures,
and achieve high accuracy on some generic recognition tasks, including {\bf SVHN}, {\bf CIFAR} and {\bf ILSVRC2012}.

The remainder of this paper is organized as follows.
Section~\ref{RelatedWork} briefly reviews related work.
The Deep Collaborative Learning (DCL) module is presented in Section~\ref{Algorithm}.
In Section~\ref{Experiments}, we evaluate DCL on a wide range of visual recognition tasks.
Conclusions are drawn in Section~\ref{Conclusions}.

\section{Related Work}
\label{RelatedWork}

\subsection{Convolutional Neural Networks}
\label{RelatedWork:CNN}

The Convolutional Neural Network (CNN) is a hierarchical model for large-scale visual recognition.
It is based on the observation that a network with enough neurons is able to fit complicated image data distribution.
Recently, the availability of large-scale training data ({\em e.g.}, ImageNet~\cite{Deng_2009_ImageNet}) and powerful GPUs
make it possible to train deep CNNs~\cite{Krizhevsky_2012_ImageNet}
which significantly outperform conventional approaches such as the Bag-of-Visual-Words (BoVW) model~\cite{Csurka_2004_Visual}.
A CNN is composed of several stacked layers.
In each of them, responses from the previous layer are convoluted with a filter bank and activated by a differentiable non-linearity.
Hence, a CNN can be considered as a composite function,
which is trained by back-propagating error signals defined by the difference between supervision and prediction at the top layer.
Efficient methods were proposed to help CNNs converge faster and prevent over-fitting,
such as ReLU activation~\cite{Krizhevsky_2012_ImageNet},
batch normalization~\cite{Ioffe_2015_Batch}, Dropout~\cite{Hinton_2012_Improving} and DisturbLabel~\cite{Xie_2016_DisturbLabel}.
It is believed that deeper networks
may produce better recognition results~\cite{Simonyan_2015_Very}\cite{Szegedy_2015_Going}\cite{He_2016_Deep}.

The intermediate responses of CNNs, {\em a.k.a.}, deep features, serve as effective image descriptions~\cite{Donahue_2014_DeCAF}.
They can be used in a wide range of computer vision tasks,
including image classification~\cite{Donahue_2014_DeCAF}\cite{Xie_2016_InterActive}\cite{Xie_2016_Towards},
image retrieval~\cite{Razavian_2014_CNN} and object detection~\cite{Girshick_2014_Rich}.
A discussion of how different CNN configurations impact deep feature performance is available in~\cite{Chatfield_2014_Return}.

\subsection{Learning Structured Visual Concepts}
\label{RelatedWork:StructuredLearning}

We aim at learning structured visual concepts, which is related to two research topics,
{\em i.e.}, part-based compositional models and component-based visual vocabulary construction.

The part-based compositional models play an important role in object recognition and detection.
It is motivated by the fact that most objects can be decomposed into some functional sub-pieces named {\em parts}.
To learn a flexible model to organize these parts,
the Deformable Part Model (DPM)~\cite{Felzenszwalb_2010_Object} optimizes an objective function
to consider both unary (appearance) terms and binary terms (spatial relationship) terms.
The detected parts are useful for object recognition,
especially in the fine-grained
scenarios~\cite{Berg_2013_POOF}\cite{Chai_2013_Symbiotic}\cite{Gavves_2013_Fine}\cite{Xie_2013_Hierarchical}.
Part-based models can be integrated into deep convolutional neural networks,
either for object recognition~\cite{Zhang_2014_Part}\cite{Xiao_2015_Application},
semantic part detection~\cite{Wang_2015_Semantic} or human pose estimation~\cite{Chen_2015_Parsing}.
There are also efforts at relating intermediate neural responses to object parts~\cite{Simon_2015_Neural}\cite{Wang_2015_Discovering}.
This work focuses on learning structured visual representation.
Compared to training a compositional model, we introduce stronger prior to facilitate explicit concept decomposition.

When there is the necessity to construct a very large visual vocabulary,
an efficient strategy is to train a set of small vocabularies, and combine them to obtain a large one.
Motivated by this idea, Product Quantization (PQ)~\cite{Jegou_2011_Product} partitions each vector into $T$ segments,
trains a small codebook on each segment independently, and approximates a vector by the concatenation of $T$ quantized codes.
To reduce quantization error, efforts are made to weaken the orthogonal constraints~\cite{Ge_2013_Optimized},
leading to composite quantization methods~\cite{Norouzi_2013_Cartesian}\cite{Zhang_2014_Composite}.
In this work, we borrow this idea to allow deep networks to simulate a large convolutional filter bank with several smaller ones.

Our work is also closely related to bilinear CNN~\cite{Lin_2015_Bilinear},
a recent model which trains pairwise discriminative features and assembles them of matrix multiplication.
This works well especially in fine-grained visual recognition, but also brings considerable computational overheads.
In comparison, our algorithm makes a reasonable assumption to reduce the computational costs significantly.

\section{Deep Collaborative Learning}
\label{Algorithm}

This section presents the {\bf Deep Collaborative Learning} (DCL) module.
The motivation is to use the combination of several small visual vocabularies to simulate the performance of a large vocabulary.
This module is especially useful in replacing a convolutional layer with a large number of filters,
{\em e.g.}, the fully-connected layers used in many network structures.

\subsection{Formulation}
\label{Algorithm:Formulation}

We start with a hidden network layer $\mathbf{X}$.
$\mathbf{X}$ is a 3D neuron cube with $W_1\times H_1\times K_1$ neurons,
where $W_1$ and $H_1$ are the width and height of the data cube, and $K_1$ is the number of channels.
We aim at producing a target layer $\mathbf{Z}$ with $W_2\times H_2\times K_2$ neurons.
This is originally implemented as a convolutional layer with $K_2$ kernels (filters).
Let ${p}\in{\mathcal{P}}$ be a spatial position in the layer $\mathbf{Z}$, where ${\left|\mathcal{P}\right|}={W_2\times H_2}$,
and ${k}\in{\left\{1,2,\ldots,K_2\right\}}$ be the index of a output channel,
the convolutional operation can be formulated as ${z_{p,k}}={\sigma\!\left(\boldsymbol{\theta}_k^\top\mathbf{x}_{p}\right)}$.
Here, $\boldsymbol{\theta}_k$ is the $k$-th filter,
and $\mathbf{x}_{p}$ is the data cube at the layer $\mathbf{X}$ corresponding to the position $p$.
$\sigma\!\left(\cdot\right)$ is the ReLU activation function~\cite{Krizhevsky_2012_ImageNet}.

Instead of constructing $K_2$ filters directly, DCL adopts a compositional strategy to perform this task.
This is motivated by some previous work~\cite{Jegou_2011_Product}\cite{Zhang_2014_Composite}
in constructing a large vocabulary for high-dimensional visual descriptors.
A DCL module consists of two stages, {\em i.e.}, branch construction and concept fusion.

At the first stage, {\em branch construction}, $T$ intermediate {\em branches} are generated.
Each of them, denoted as $\mathbf{Y}^{\left(t\right)}$, has $W_2\times H_2\times M^{\left(t\right)}$ neurons, ${t}={1,2,\ldots,T}$.
For simplicity, we assume these branches have the same spatial resolution as the original output layer $\mathbf{Z}$,
{\em i.e.}, the set $\mathcal{P}$ is shared among $\mathbf{Z}$ and all $\mathbf{Y}^{\left(t\right)}$'s.
This is easily implemented by using the same convolutional kernel size and spatial stride.
Following the above definitions, the convolutional operation can be denoted as
${y_{p,m}^{\left(t\right)}}={\boldsymbol{\theta}_m^{\left(t\right)\top}\mathbf{x}_{p}}$, ${m}={1,2,\ldots,M^{\left(t\right)}}$
After this stage, at each positions $p$,
we obtain $T$ vectors, and the $t$-th of them is ${\mathbf{y}_p^{\left(t\right)}}\in{\mathbb{R}^{M^{\left(t\right)}}}$.

At the second stage, {\em concept fusion}, we combine $T$ vectors together at each position $p$ individually.
Recall that DCL is a replacement of the original convolutional layer,
therefore we simply keep the number of output channels, {\em i.e.}, $K_2$, unchanged.
Thus, we need a mapping function $\mathbf{f}:\mathbb{R}^{{\sum_t}M^{\left(t\right)}}\rightarrow\mathbb{R}^{K_2}$.
In practice, this is implemented by fully-connecting each ${\mathbf{y}_p^{\left(t\right)}}\in{\mathbb{R}^{M^{\left(t\right)}}}$
to a $K_2$-dimensional vector ${\mathbf{v}_p^{\left(t\right)}}\in{\mathbb{R}^{K_2}}$,
{\em i.e.}, ${\mathbf{v}_p^{\left(t\right)}}={\sigma\!\left(\mathbf{W}^{\left(t\right)\top}\mathbf{y}_p^{\left(t\right)}\right)}$,
and performing an element-wise multiplication followed by $T$-th root to fuse all $\mathbf{v}_p^{\left(t\right)}$'s together,
{\em i.e.}, ${z_{p,k}}={\sqrt[T]{{\prod_t}v_{p,k}^{\left(t\right)}+\varepsilon}}$.
Here, $\mathbf{W}^{\left(t\right)}$ is a weighting matrix with $M^{\left(t\right)}\times K_2$ elements,
and ${\varepsilon}={10^{-T}}$ is a small floating point number to avoid numerical instability in gradient computation.

\newcommand{\figurewidth}{8.0cm}
\begin{figure}
\begin{center}
    \includegraphics[width=\figurewidth]{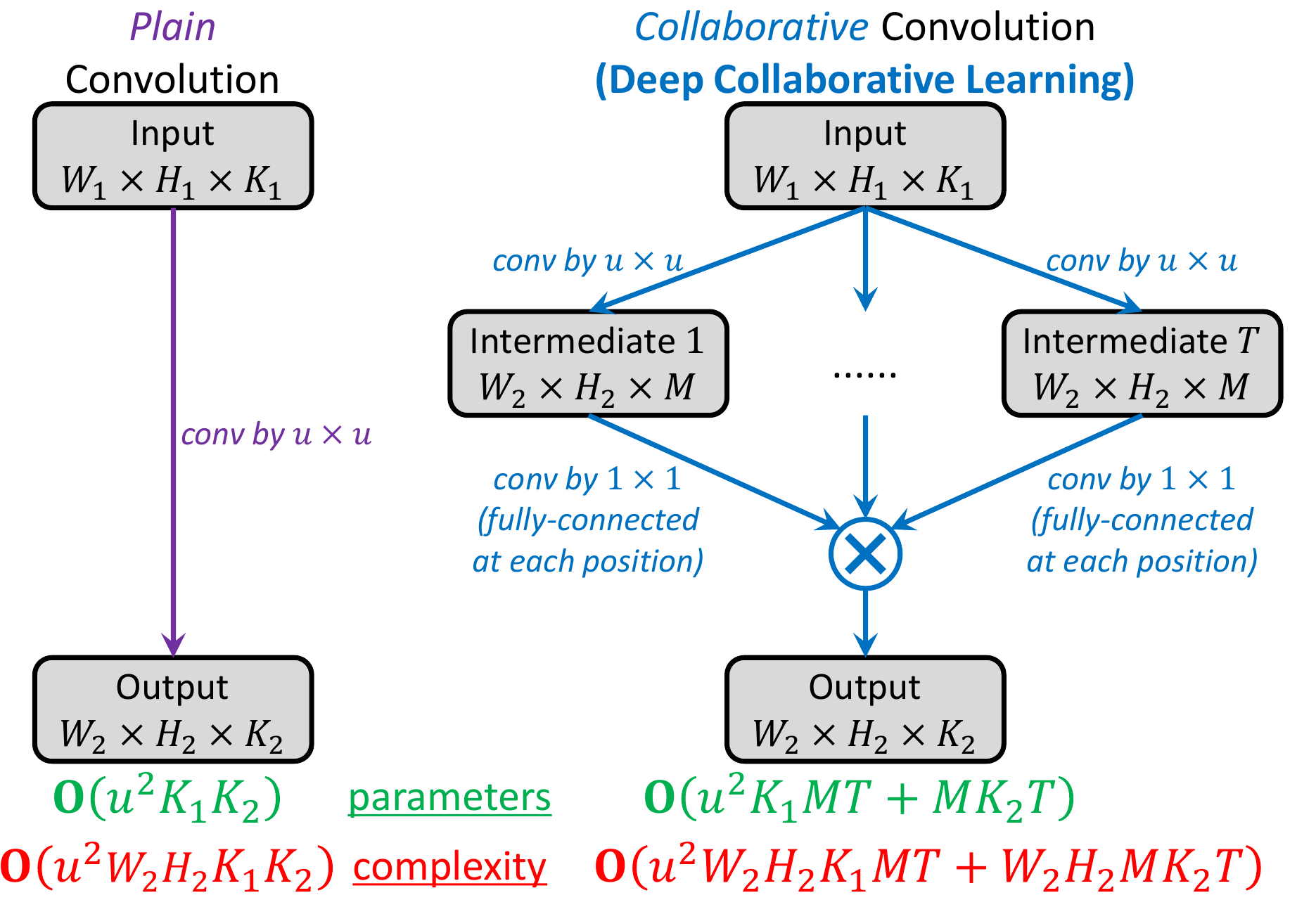}
\end{center}
\caption{
    Illustration of the original convolution (left) and the proposed Deep Collaborative Learning (DCL) module (right).
    We also compare the number of parameters and computational complexity of both models.
    For simplicity, we assume all branches have the same number of filters $M$.
}
\label{Fig:DeepCollaborativeLearning}
\end{figure}

The comparison between an original convolutional layer and a DCL module is illustrated in Figure~\ref{Fig:DeepCollaborativeLearning}.
The number of branches $T$ and the number of filters in each branch $M^{\left(t\right)}$ are hyper-parameters,
which need to be investigated in later experiments (see Section~\ref{Experiments:Structured}).

\subsection{Learning Compositional Visual Concepts}
\label{Algorithm:Compositional}

We show that DCL is able to learn visual concepts in a compositional manner.
We investigate an output unit ${z_{p,k}}>{0}$, so that for all ${t}={1,2,\ldots,T}$,
${\sigma\!\left({\sum_{m^{\left(t\right)}=1}^{M^{\left(t\right)}}}W_{k,i}^{\left(1\right)}y_{p,i}^{\left(1\right)}\right)}>{0}$,
thus ${\sigma\!\left({\sum_{m^{\left(t\right)}=1}^{M^{\left(t\right)}}}W_{k,i}^{\left(1\right)}y_{p,i}^{\left(1\right)}\right)}
    ={{\sum_{m^{\left(t\right)}=1}^{M^{\left(t\right)}}}W_{k,i}^{\left(1\right)}y_{p,i}^{\left(1\right)}}$.
Following the formulation and ignoring the $\varepsilon$ term, we have:
\begin{eqnarray*}
{z_{p,k}^T} & = & {{\prod_{t=1}^T}
        \left({\sum_{m^{\left(t\right)}=1}^{M^{\left(t\right)}}}W_{k,i}^{\left(t\right)}y_{p,i}^{\left(t\right)}\right)} \\
    & = & {{\sum_{m^{\left(1\right)}=1}^{M^{\left(1\right)}}}\cdots{\sum_{m^{\left(T\right)}=1}^{M^{\left(T\right)}}}
        {\prod_{t=1}^T}W_{k,m^{\left(t\right)}}^{\left(t\right)}{\prod_{t=1}^T}y_{p,m^{\left(t\right)}}^{\left(t\right)}} \\
    & = & {{\sum_{m^{\left(1\right)}=1}^{M^{\left(1\right)}}}\cdots{\sum_{m^{\left(T\right)}=1}^{M^{\left(T\right)}}}
        {\prod_{t=1}^T}W_{k,m^{\left(t\right)}}^{\left(t\right)}
        {\prod_{t=1}^T}\sigma\!\left(\boldsymbol{\theta}_{m^{\left(t\right)}}^{\left(t\right)\top}\mathbf{x}_p\right)}.
\end{eqnarray*}
We focus on the last term ${\prod_t}\sigma\!\left(\boldsymbol{\theta}_{m^{\left(t\right)}}^{\left(t\right)\top}\mathbf{x}_p\right)$,
which is the multiplication of $T$ convolutional results on the same input patch $\mathbf{x}_p$.
To obtain a positive value, each $\boldsymbol{\theta}_{m^{\left(t\right)}}^{\left(t\right)\top}\mathbf{x}_p$ should be positive,
which means that all $T$ filters are fired at position $p$.
Note that the above formula enumerates all ${\prod_t}M^{\left(t\right)}$ combinations of $\left(m^{\left(t\right)}\right)_{t=1}^T$,
or equivalently, we consider ${\prod_t}M^{\left(t\right)}$ compositional filters.
Although this number grows exponentially with $T$, the model complexity merely increases linearly.

\subsection{Relationship to Other Work}
\label{Algorithm:Relationship}

DCL is closely related to the Bilinear-CNN (BCNN) model~\cite{Lin_2015_Bilinear}.
Both models are motivated by the need of integrating different sources of visual features.
To directly compare to BCNN, we set ${T}={2}$ in DCL.
Note that BCNN also specifies a set $\mathcal{P}$ of spatial positions and uses two pre-trained networks for feature extraction.
The difference lies in the method of feature fusion, {\em i.e.}, the {\em concept combination} stage in DCL.

At each spatial position ${p}\in{\mathcal{P}}$, BCNN extracts two types of features
${\mathbf{y}_p^{\left(1\right)}}\in{\mathbb{R}^{M^{\left(1\right)}}}$ and
${\mathbf{y}_p^{\left(2\right)}}\in{\mathbb{R}^{M^{\left(2\right)}}}$, respectively,
and computes the outer-product ${\mathbf{y}_p^{\left(1\right)}\otimes\mathbf{y}_p^{\left(2\right)}}\in
    {\mathbb{R}^{M^{\left(1\right)}\cdot M^{\left(2\right)}}}$.
Then, the product is fully-connected to a vector ${\mathbf{z}_p}\in{\mathbb{R}^{K_2}}$ at the next stage
with a weight matrix ${\mathbf{U}}\in{\mathbb{R}^{M^{\left(1\right)}\cdot M^{\left(2\right)} \cdot K_2}}$.
For an element $z_{p,k}$ in $\mathbf{z}_p$,
its value is defined as ${z_{p,k}}={{\sum_{i,j}}U_{i,j,k}y_{p,i}^{\left(1\right)}y_{p,j}^{\left(2\right)}}$.
DCL works in a different manner.
After ${\mathbf{y}_p^{\left(1\right)}}\in{\mathbb{R}^{M^{\left(1\right)}}}$ and
${\mathbf{y}_p^{\left(2\right)}}\in{\mathbb{R}^{M^{\left(2\right)}}}$ are computed,
they are first fully-connected to the next stage with two weight matrices
${\mathbf{W}^{\left(1\right)}}\in{\mathbb{R}^{M^{\left(1\right)}\cdot K_2}}$
and ${\mathbf{W}^{\left(2\right)}}\in{\mathbb{R}^{M^{\left(2\right)}\cdot K_2}}$,
then fused via element-wise multiplication followed by taking the square root.
Hence, an element $z_{p,k}$ in the vector $\mathbf{z}_p$ takes the form
${z_{p,k}}={\sqrt{\left({\sum_i}W_{p,i}^{\left(1\right)}y_{p,i}^{\left(1\right)}\right)\cdot
    \left({\sum_j}W_{p,j}^{\left(2\right)}y_{p,j}^{\left(2\right)}\right)}}$.

If ${U_{p,i,j}}={W_{p,i}^{\left(1\right)}\times W_{p,j}^{\left(2\right)}}$ holds for any $\left(i,j\right)$,
we have ${\sum_{i,j}}U_{p,i,j}x_{p,i}^{\left(1\right)}x_{p,j}^{\left(2\right)}=
\left({\sum_i}W_{p,i}^{\left(1\right)}x_{p,i}^{\left(1\right)}\right)\cdot
    \left({\sum_j}W_{p,j}^{\left(2\right)}x_{p,j}^{\left(2\right)}\right)$.
This means that DCL is a constrained case of BCNN, which assumes the decomposable property of the weights.
To illustrate this, we explain the neural responses at the position $p$,
{\em i.e.}, $\mathbf{x}_p^{\left(1\right)}$ and $\mathbf{x}_p^{\left(2\right)}$, as some types of visual attributes.
As an example, let $\mathbf{x}_p^{\left(1\right)}$ and $\mathbf{x}_p^{\left(2\right)}$
represent color and shape features at the given position, respectively.
Consider two colors, {\em red} and {\em blue}, and two shapes, {\em triangle} and {\em square},
and their combination produces ${2\times2}={4}$ compound visual concepts.
Using BCNN, these four compound concepts may be assigned independent weights.
Using DCL, on the other hand, the weights are constrained,
{\em e.g.}, if a {\em red triangle} is $50\%$ more important than a {\em red square}
({\em i.e.}, the weight on a {\em triangle} is $50\%$ higher than the weight on a {\em square}),
then a {\em blue triangle} is also $50\%$ more important than a {\em blue square}.
Although this constraint applies to each single filter, the ratio can vary from filter to filter,
{\em e.g.}, in another filter, the weight on a {\em triangle} may be $30\%$ lower than the weight on a {\em square}.
Such an assumption is reasonable, since each filter often focuses on a specific combination of visual attributes,
and the preference within one attribute is often independent to the preference of other attributes.
This assumption, on the other hand, brings the benefit of a reduced number of parameters.
For each of the $K_2$ filters, BCNN requires $M^{\left(1\right)}\cdot M^{\left(2\right)}$ weights,
while DCL only needs $M^{\left(1\right)}+M^{\left(2\right)}$.
In our experiments, $M^{\left(1\right)}$ and $M^{\left(2\right)}$ are always large ({\em e.g.}, tens or hundreds),
thus using DCL leads to a less complicated model and, consequently, less risk of over-fitting.

In another perspective, DCL increases the depth of the network but decreases the number of parameters.
This is achieved by decomposing the visual vocabulary in the channel domain.
A similar effort is made by {\bf VGGNet}~\cite{Simonyan_2015_Very},
which uses two consecutive $3\times3$ layers to simulate the performance of a $5\times5$ layer.
This is to decompose the vocabulary in the spatial domain.

\subsection{Computational Complexity}
\label{Algorithm:Complexity}

We analyze the number of trainable parameters and the computational complexity of the original convolutional layer and DCL.
Denote $u\times u$ as the kernel size used in original convolution (${\mathbf{X}}\Rightarrow{\mathbf{Z}}$)
and each of the intermediate branches (${\mathbf{X}}\Rightarrow{\mathbf{Y}^{\left(t\right)}}$, ${t}={1,2,\ldots T}$).
An original convolutional layer requires $u^2K_1K_2$ parameters, and $O\!\left(u^2W_2H_2K_1K_2\right)$ complexity,
while a DCL module requires $u^2K_1{\sum_t}M^{\left(t\right)}+K_2{\sum_t}M^{\left(t\right)}$ parameters,
and $O\!\left(u^2W_2H_2K_1{\sum_t}M^{\left(t\right)}+W_2H_2K_2{\sum_t}M^{\left(t\right)}\right)$ complexity.
Note that ${u^2K_1{\sum_t}M^{\left(t\right)}+K_2{\sum_t}M^{\left(t\right)}}\leqslant{u^2K_1K_2}$
is equivalent to ${u^2W_2H_2K_1{\sum_t}M^{\left(t\right)}+W_2H_2K_2{\sum_t}M^{\left(t\right)}}\leqslant{u^2W_2H_2K_1K_2}$.

We discuss the above inequality for some special cases.
If ${u}\geqslant{3}$ and ${K_1}\approx{K_2}$ (a common setting before the fully-connected layers),
then ${u^2K_1}\gg{K_2}$, and so we can ignore the second term in the left-hand side yielding ${{\sum_t}M^{\left(t\right)}}<{K_2}$.
If ${u}={1}$ (a regular case between fully-connected layers where the spatial resolution of convolution is $1\times1$),
the inequality becomes ${\left(K_1+K_2\right){\sum_t}M^{\left(t\right)}}\leqslant{K_1K_2}$.
If we further have ${K_1}={K_2}$, then ${2\times{\sum_t}M^{\left(t\right)}}\leqslant{K_1}={K_2}$.
In experiments, we always set ${{\sum_t}M^{\left(t\right)}}\leqslant{K_2/2}$ to guarantee reduced complexity.

\subsection{Training a Multi-Branch Model}
\label{Algorithm:MultiBranch}

In training a DCL model with more than two intermediate branches (${T}>{2}$),
the high-order root operation may cause instable numerical issues.
To deal with this, we suggest a {\em stochastic} training strategy, which is opposite to the original {\em deterministic} structure.
In each training iteration, we randomly activate $2$ out of $T$ branches, and temporarily disable other branches.
This is to say, we only allow the parameters in the active branches to be trained in each iteration.
In the testing phase, we compute the expectation of such random selection.
This is done by enumerating all ${T\choose2}$ branch pairs and computing the averaged neural responses over all these choices.

Besides the numerical issue, the stochastic strategy brings two benefits.
First, the network structure used in each training iteration is different, which helps prevent over-fitting.
Second, as $T$ goes up, it becomes more difficult for all $T$ filters to fire at a spatial position.
The stochastic strategy considers a pair of filters at each time,
possibly giving a neuron positive response when $2$ out of $T$ filters are fired.
The effectiveness of this strategy is verified in experiments (see Section~\ref{Experiments:Structured:Comparison}).

\section{Experiments}
\label{Experiments}

\subsection{Multi-Digit Number Classification}
\label{Experiments:Structured}

We first evaluate DCL on multi-digit number classification to show its ability in discovering feature co-occurrence.

\subsubsection{Dataset Construction}
\label{Experiments:Structured:Dataset}

\newcommand{\bigfigurewidth}{17.0cm}
\begin{figure*}
\begin{center}
    \includegraphics[width=\bigfigurewidth]{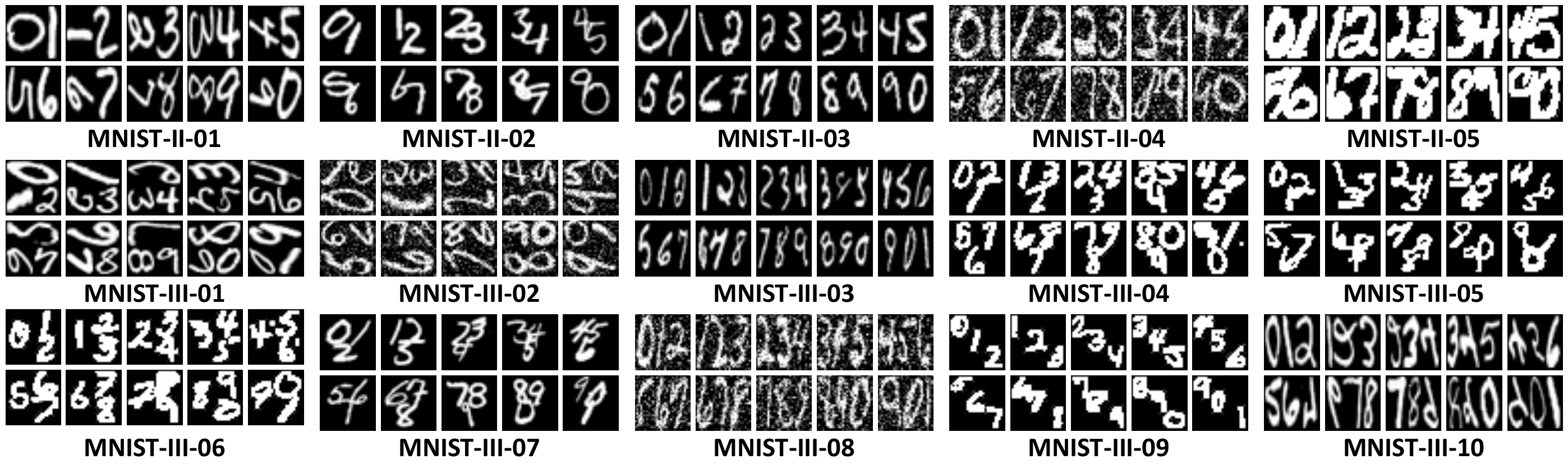}
\end{center}
\caption{
    Configuration and sample images for $5$ two-digit recognition and $10$ three-digit datasets.
    A digit may undergo various types of variations, including scale variance, slight rotation and flipping.
    Additional noise may also be added.
    A larger dataset ID implies a higher level of difficulty (see Table~\ref{Tab:StructuredDatasets} for reference).
}
\label{Fig:MultiDigitDatasets}
\end{figure*}

We construct $5$ two-digit and $10$ three-digit datasets based on the basic {\bf MNIST}~\cite{LeCun_1998_Gradient}.
These datasets differ from each other in many aspects, including the central position of each digit,
the scaling, rotation and flipping properties of each digit, and if additional noise is added to the image.
We generate the training and testing subsets of each dataset using only the training and testing data of {\bf MNIST}.
Although the number of categories increases significantly ($100$ for two-digit set and $1\rm{,}000$ for three-digit sets),
we keep the amount of training ($60\rm{,}000$) and testing ($10\rm{,}000$) images unchanged.
This increases the risk of over-fitting for the network models with large numbers of parameters.

The configuration and example images of these multi-digit datasets are shown in Figure~\ref{Fig:MultiDigitDatasets}.
When we need to generate a number, we first randomly sample each of its digit from the given (training or testing) set,
segment them from the original image, and assemble them into the target image.
The geometric property of each digit ({\em i.e.}, central position, scaling, rotation and flipping)
are determined by the fixed configuration of each dataset but may undergo slight random variation.
In some cases, neighboring digits may overlap with each other, which increases the difficulty of recognition.
Finally, we compute the minimal bounding box of the multi-digit number,
and rescale it into $28\times28$ (the same size as the original {\bf MNIST} dataset).

\subsubsection{Baseline and Oracle}
\label{Experiments:Structured:Comparison}

We use {\bf LeNet}~\cite{LeCun_1990_Handwritten} as our baseline.
This is a shallow network with two convolutional layers, two pooling layers and two fully-connected layers.
The network architecture can be abbreviated as:
\begin{spverbatim}
C5@20-MP2S2-C5@50-MP2S2-FC500-D0.5-OUT.
\end{spverbatim}
\noindent
Here, {\tt C5@20} is a convolutional layer with a kernel size $5$, a default spatial stride $1$ and the number of kernels $20$;
{\tt MP2S2} is a max-pooling layer with a kernel size $2$ and a spatial stride $2$,
{\tt FC500} is a fully-connected layer with $500$ outputs, and {\tt D0.5} is a Dropout layer with a drop ratio $0.5$.
{\tt OUT} is the output layer, {\em i.e.}, a fully-connected layer with a proper number ($100$ or $1\rm{,}000$) of filters.

We try different model options and parameters,
{\em i.e.}, replacing different fully-connected layers (the first and/or the second) with DCL,
constructing different numbers of branches (two or three), and, in the case of three branches,
using either the deterministic or stochastic training strategy as described in Section~\ref{Algorithm:MultiBranch}.
Each branch in DCL contains $1/5$ of the original number of filters.
We denote the first and second fully-connected layers as {\bf A} and {\bf B}, respectively,
and use {\bf 2}, {\bf 3D} and {\bf 3S} to represent training two branches or three branches with different strategies.
To summarize, the six DCL models can be denoted as {\bf DCL-A2}, {\bf DCL-A3D}, {\bf DCL-A3S},
{\bf DCL-B2}, {\bf DCL-B3D} and {\bf DCL-B3S}, respectively.

We also compare DCL with the so-called {\em oracle} algorithm,
which works by training two or three sub-classifiers, each of which for one digit individually,
and combining their prediction as the final classification result.
Note that this classifier uses strong prior knowledge (each number consists of into several digits),
and designs a specific scheme which cannot be generalized to other classification problems.
Although it is not fair in direct comparison with DCL, the oracle algorithm provides an upper-bound of the recognition accuracy.

\newcommand{\colwidthA}{1.45cm}
\newcommand{\colwidthB}{1.35cm}
\begin{table*}
\centering
\begin{tabular}{|l||R{\colwidthA}|R{\colwidthB}||R{\colwidthB}|R{\colwidthB}|
                    R{\colwidthB}||R{\colwidthA}|R{\colwidthB}|R{\colwidthA}|}
\hline
\multirow{2}{*}{Dataset} & \multirow{2}{*}{Baseline}                   & \multirow{2}{*}{Oracle}
                         & \multicolumn{6}{c|}{Deep Collaborative Learning ({\bf DCL})}                              \\
\cline{4-9}
{}                       & {}                   & {}                   & {\bf A2}             & {\bf A3S}
                         & {\bf A3D}            & {\bf B2}             & {\bf B3S}            & {\bf B3D}            \\
\hline\hline
{\bf MNIST-II-01}        & $ 1.96(03)$          & $\mathbf{ 1.07}(04)$ & $ 1.76(03)$          & $\mathbf{ 1.73}(04)$
                         & $ 1.85(09)$          & $ 2.15(06)$          & $ 2.11(07)$          & $ 2.12(12)$          \\
\hline
{\bf MNIST-II-02}        & $ 2.58(07)$          & $\mathbf{ 2.02}(11)$ & $ 2.38(05)$          & $\mathbf{ 2.36}(06)$
                         & $ 2.40(10)$          & $ 2.56(06)$          & $ 2.46(04)$          & $ 2.50(05)$          \\
\hline
{\bf MNIST-II-03}        & $ 2.99(07)$          & $\mathbf{ 2.12}(15)$ & $ 2.80(06)$          & $\mathbf{ 2.76}(14)$
                         & $\mathbf{ 2.76}(05)$ & $ 3.17(15)$          & $ 3.09(05)$          & $ 3.08(15)$          \\
\hline
{\bf MNIST-II-04}        & $ 3.55(11)$          & $\mathbf{ 2.85}(09)$ & $\mathbf{ 3.31}(10)$ & $\mathbf{ 3.31}(06)$
                         & $ 3.44(06)$          & $ 3.67(15)$          & $ 3.44(02)$          & $ 3.54(11)$          \\
\hline
{\bf MNIST-II-05}        & $ 5.10(14)$          & $\mathbf{ 4.23}(06)$ & $\mathbf{ 4.86}(15)$ & $\mathbf{ 4.86}(13)$
                         & $ 4.94(14)$          & $ 5.86(17)$          & $ 5.81(05)$          & $ 5.91(18)$          \\
\hline\hline
{\bf MNIST-III-01}       & $ 3.08(14)$          & $\mathbf{ 2.15}(08)$ & $ 2.82(07)$          & $\mathbf{ 2.71}(06)$
                         & $ 2.82(05)$          & $ 2.96(14)$          & $ 2.91(08)$          & $ 3.00(10)$          \\
\hline
{\bf MNIST-III-02}       & $ 5.38(16)$          & $\mathbf{ 3.39}(09)$ & $ 4.73(15)$          & $\mathbf{ 4.54}(09)$
                         & $ 4.81(05)$          & $ 5.09(08)$          & $ 5.08(05)$          & $ 5.46(22)$          \\
\hline
{\bf MNIST-III-03}       & $ 6.30(13)$          & $\mathbf{ 4.93}(18)$ & $ 5.76(12)$          & $ 5.73(08)$
                         & $\mathbf{ 5.72}(05)$ & $ 6.14(17)$          & $ 6.31(12)$          & $ 6.48(10)$          \\
\hline
{\bf MNIST-III-04}       & $ 7.81(16)$          & $\mathbf{ 5.64}(22)$ & $ 7.22(11)$          & $ 7.19(05)$
                         & $ 7.25(16)$          & $ 5.75(17)$          & $\mathbf{ 5.65}(16)$ & $ 5.79(24)$          \\
\hline
{\bf MNIST-III-05}       & $ 7.88(20)$          & $\mathbf{ 7.08}(15)$ & $ 7.33(11)$          & $\mathbf{ 7.20}(19)$
                         & $ 7.39(10)$          & $ 7.71(09)$          & $ 7.64(13)$          & $ 7.98(21)$          \\
\hline
{\bf MNIST-III-06}       & $ 8.03(16)$          & $\mathbf{ 5.52}(11)$ & $ 7.31(12)$          & $ 7.08(20)$
                         & $ 7.26(07)$          & $\mathbf{ 5.52}(20)$ & $ 5.63(05)$          & $ 5.79(28)$          \\
\hline
{\bf MNIST-III-07}       & $ 8.98(30)$          & $\mathbf{ 7.59}(22)$ & $ 8.41(25)$          & $\mathbf{ 8.38}(14)$
                         & $ 8.47(34)$          & $ 8.87(30)$          & $ 8.95(21)$          & $ 9.09(28)$          \\
\hline
{\bf MNIST-III-08}       & $ 9.57(19)$          & $\mathbf{ 8.22}(16)$ & $ 8.94(19)$          & $\mathbf{ 8.88}(11)$
                         & $ 8.94(08)$          & $ 9.60(17)$          & $ 9.49(14)$          & $ 9.79(20)$          \\
\hline
{\bf MNIST-III-09}       & $10.08(27)$          & $\mathbf{ 9.34}(15)$ & $ 9.86(17)$          & $\mathbf{ 9.71}(16)$
                         & $ 9.90(14)$          & $10.07(43)$          & $ 9.93(08)$          & $10.03(28)$          \\
\hline
{\bf MNIST-III-10}       & $10.12(25)$          & $\mathbf{ 7.61}(12)$ & $ 9.17(12)$          & $\mathbf{ 9.14}(07)$
                         & $ 9.19(07)$          & $ 9.74(27)$          & $ 9.97(10)$          & $10.11(31)$          \\
\hline
\end{tabular}
\caption{
    Classification error rates ($\%$) on the multi-digit datasets.
    Please refer to the texts for detailed model configurations.
}
\label{Tab:StructuredDatasets}
\end{table*}

Results are summarized in Table~\ref{Tab:StructuredDatasets}.
First, we observe that the oracle algorithm produces much higher classification accuracy than the baseline model.
The benefit mainly comes from the extra knowledge, which decomposes the complicated problem into several sub-classifiers,
each of which only needs to distinguish $10$ classes, reducing the recognition difficulty significantly.
However, such a method cannot be applied to generic classification problems.
DCL, on the other hand, does not assume and rely on any extra information,
but only designs a compositional structure to facilitate the network to discover feature co-occurrence.
Adding DCL on the first fully-connected layer consistently improves the baseline performance on every dataset.
In some situations, {\em e.g.}, {\bf MNIST-III-05}, the best DCL model is even comparable to the oracle.

\subsubsection{Parameters and Complexity}
\label{Experiments:Structured:Parameters}

We discuss on the impact of some model options.
\begin{itemize}
\item {\bf The layer replaced by DCL.}
We find that adding DCL on the first fully-connected layer always outperforms the baseline.
Adding DCL on the second fully-connected layer makes the model instable,
{\em i.e.}, sometimes it is significantly better ({\em e.g.}, {\bf MNIST-III-06}),
sometimes it is even worse than the baseline ({\em e.g.}, {\bf MNIST-II-05}).
Motivated by this, we do not try to replace two fully-connected layers simultaneously.
\item {\bf The number of branches and the training strategy.}
On all three-digit datasets, the three-branch models work better than the two-branches models significantly.
However, the advantage becomes much smaller when the models are evaluated on two-digit datasets.
This suggests that the complexity of the designed structured model should be related to the difficulty of the dataset.
We can certainly design a over-complicated model to deal with a simple task,
but the increasing number of parameters may incur over-fitting (see later experiments).
This is the reason why we do not train models with more than three branches.
On the other hand, the stochastic training strategy often works better than the deterministic strategy,
which works by randomly switching off branches and reducing the number of parameters in each iteration.
\item {\bf The number of filters in each branch.}
We evaluate the {\bf DCL-A2} model on the most difficult three-digit dataset ({\bf MNIST-III-10}).
Results with respect to different parameters are shown in Figure~\ref{Fig:Filters}.
We can see that the classification accuracy goes up with the increasing amount of filters.
However, a large filter bank does not help much in recognition meanwhile brings heavier computational overheads.
\end{itemize}
In the later experiments on generic classification, we will preserve the best options learned here,
{\em i.e.}, replacing the {\em first} fully-connected layer with DCL,
using three branches (as natural image often contains complicated situations) with the stochastic training strategy.

\newcommand{\subfigurewidth}{8.0cm}
\newcommand{\scatterwidth}{6.5cm}
\newcommand{\subfigurekern}{0.5cm}
\begin{figure*}
\begin{center}
\begin{minipage}{\subfigurewidth}
\centering
    \includegraphics[width=\scatterwidth]{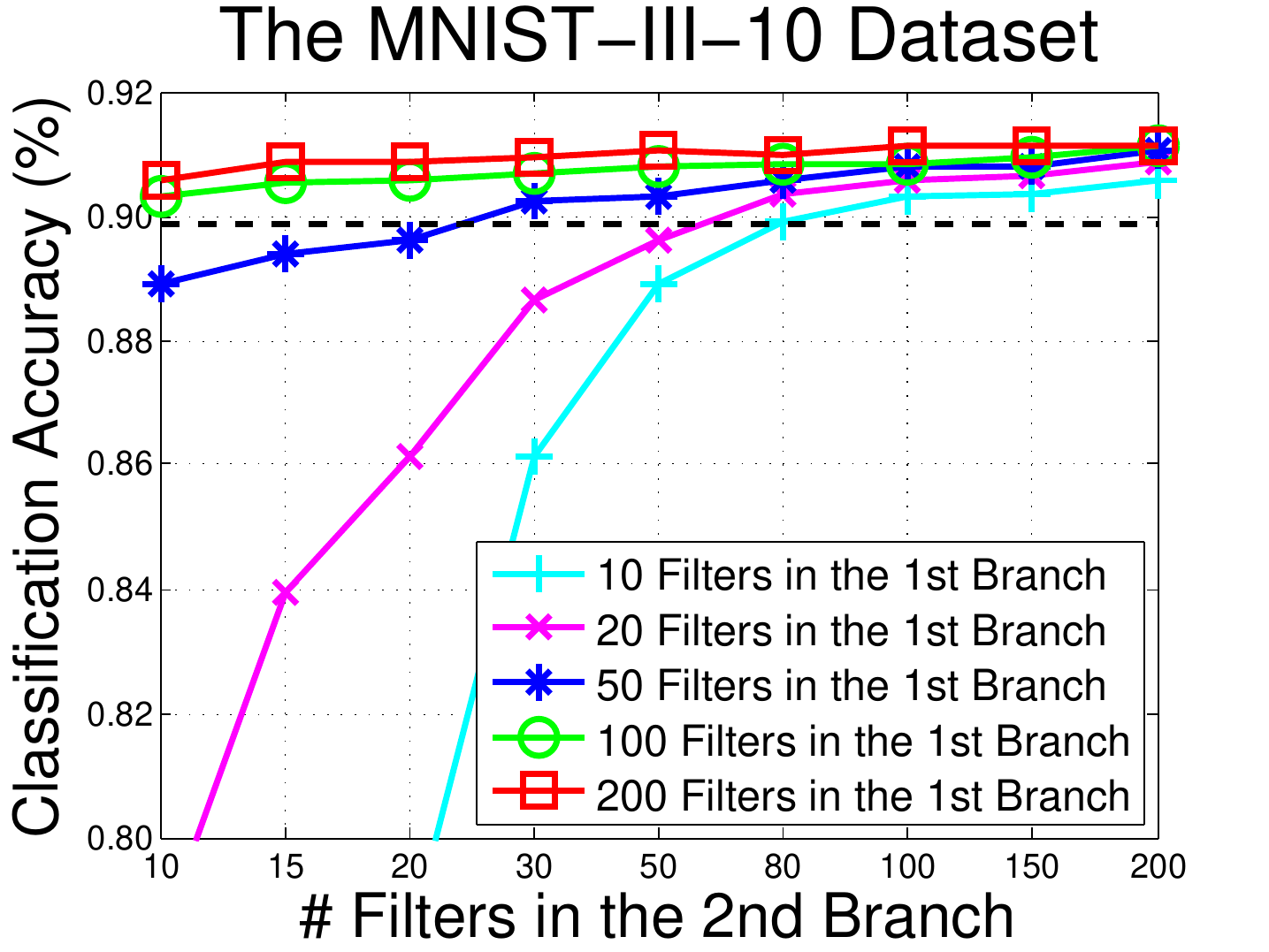}
\caption{
    Recognition accuracy ($\%$) on {\bf MNIST-III-10} with different numbers of filters in the {\bf DCL-A2} model.
    The black dashed line indicates the baseline accuracy.
}
\label{Fig:Filters}
\end{minipage}
\hspace{\subfigurekern}
\begin{minipage}{\subfigurewidth}
\centering
    \includegraphics[width=\scatterwidth]{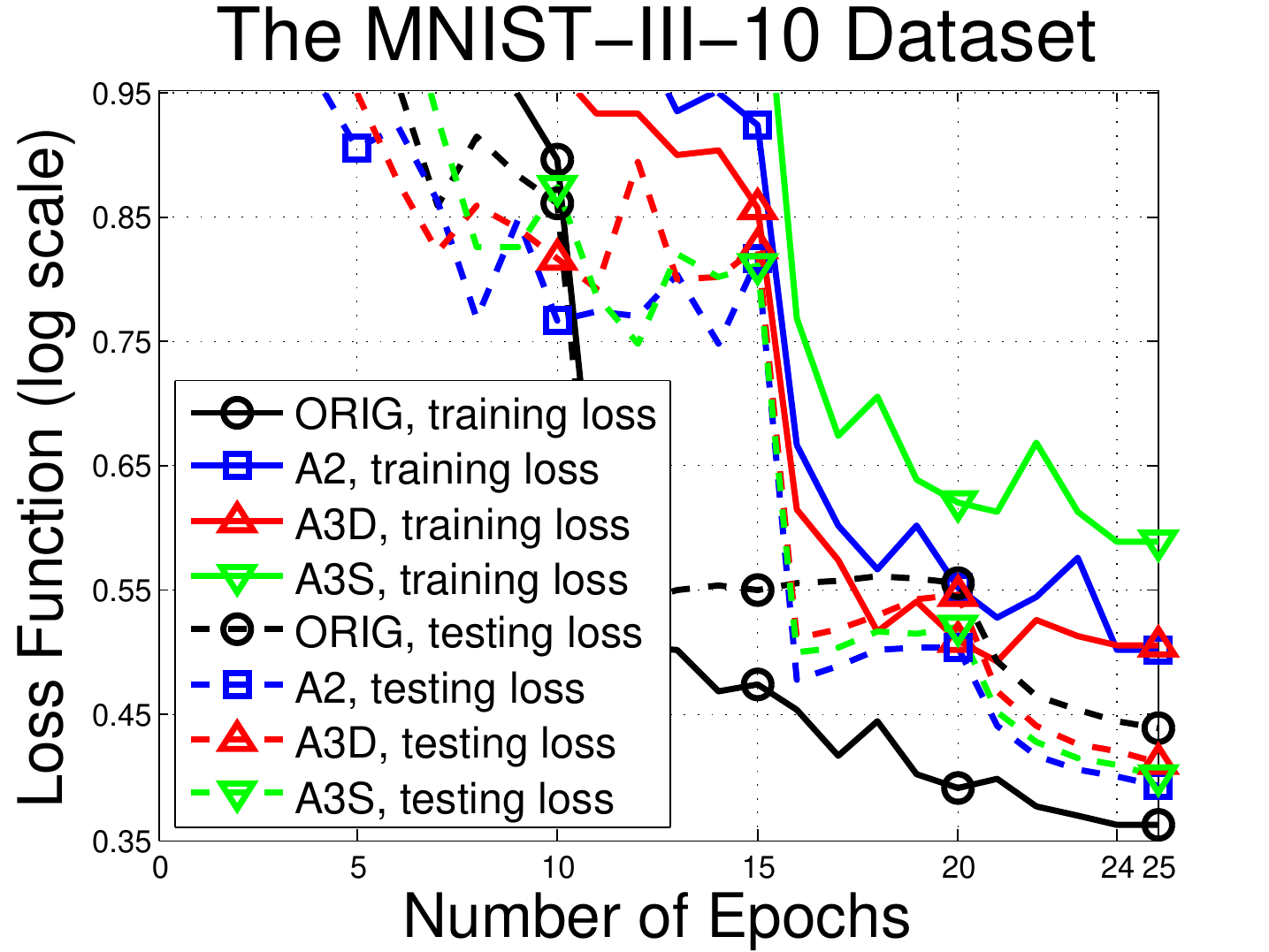}
\caption{
    Training and testing loss curves on {\bf MNIST-III-10} using the baseline and three different DCL models,
    {\em i.e.}, {\bf DCL-A2}, {\bf DCL-A3D} and {\bf DCL-A3S}.
}
\label{Fig:Curves}
\end{minipage}
\end{center}
\end{figure*}

Finally, on the most challenging {\bf MNIST-III-10} dataset,
we plot the training and testing curves of the baseline and three DCL models in Figure~\ref{Fig:Curves}.
Not surprisingly, by reducing the number of parameters,
DCL largely alleviates the over-fitting phenomenon in the training process.
This is especially useful when the amount of training data is limited.

\subsubsection{Learning Complementary Visual Knowledge}
\label{Experiments:Structured:ComplementaryKnowledge}

\renewcommand{\figurewidth}{8.0cm}
\begin{figure*}
\begin{center}
    \includegraphics[width=\figurewidth]{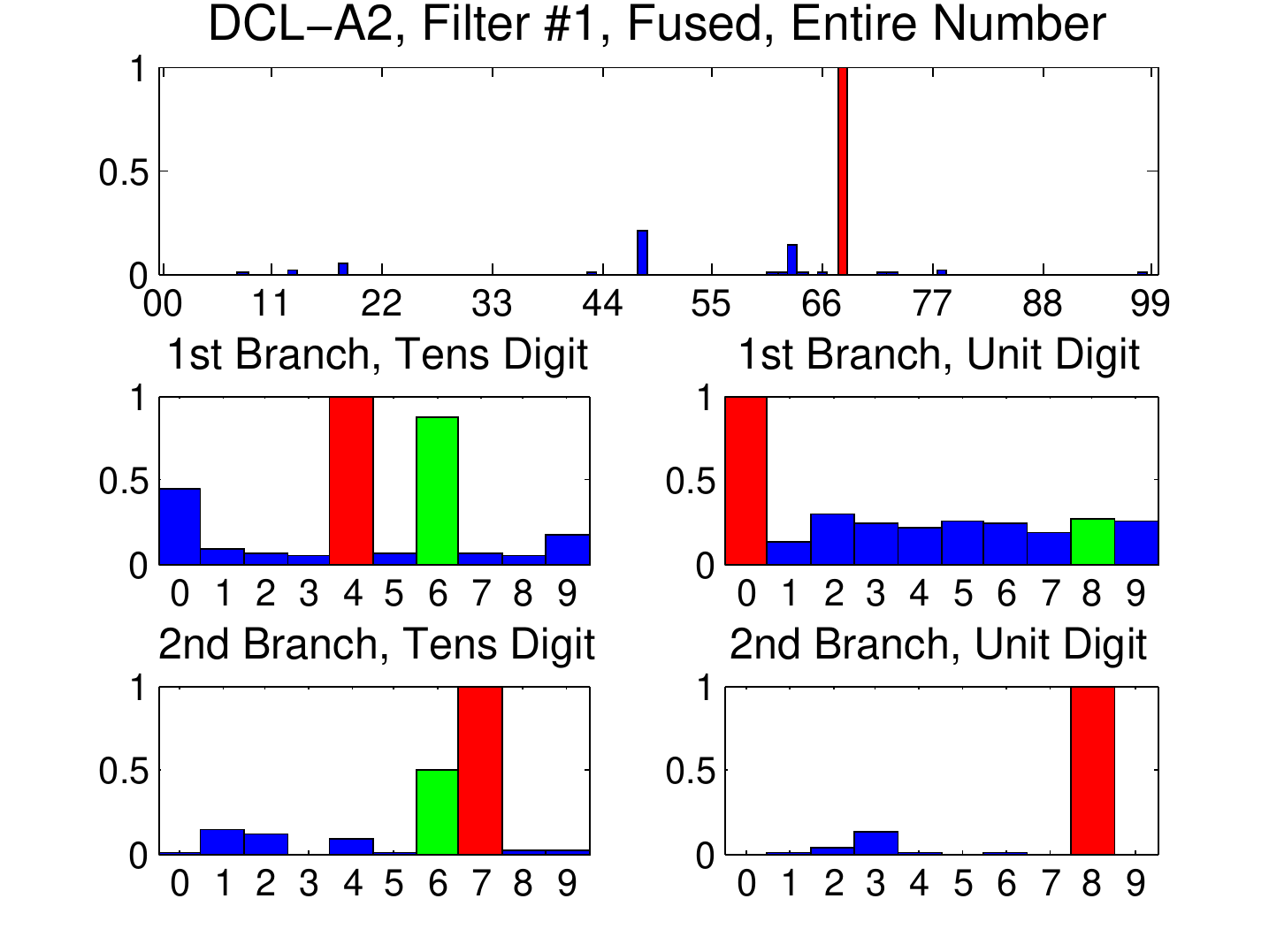}
    \includegraphics[width=\figurewidth]{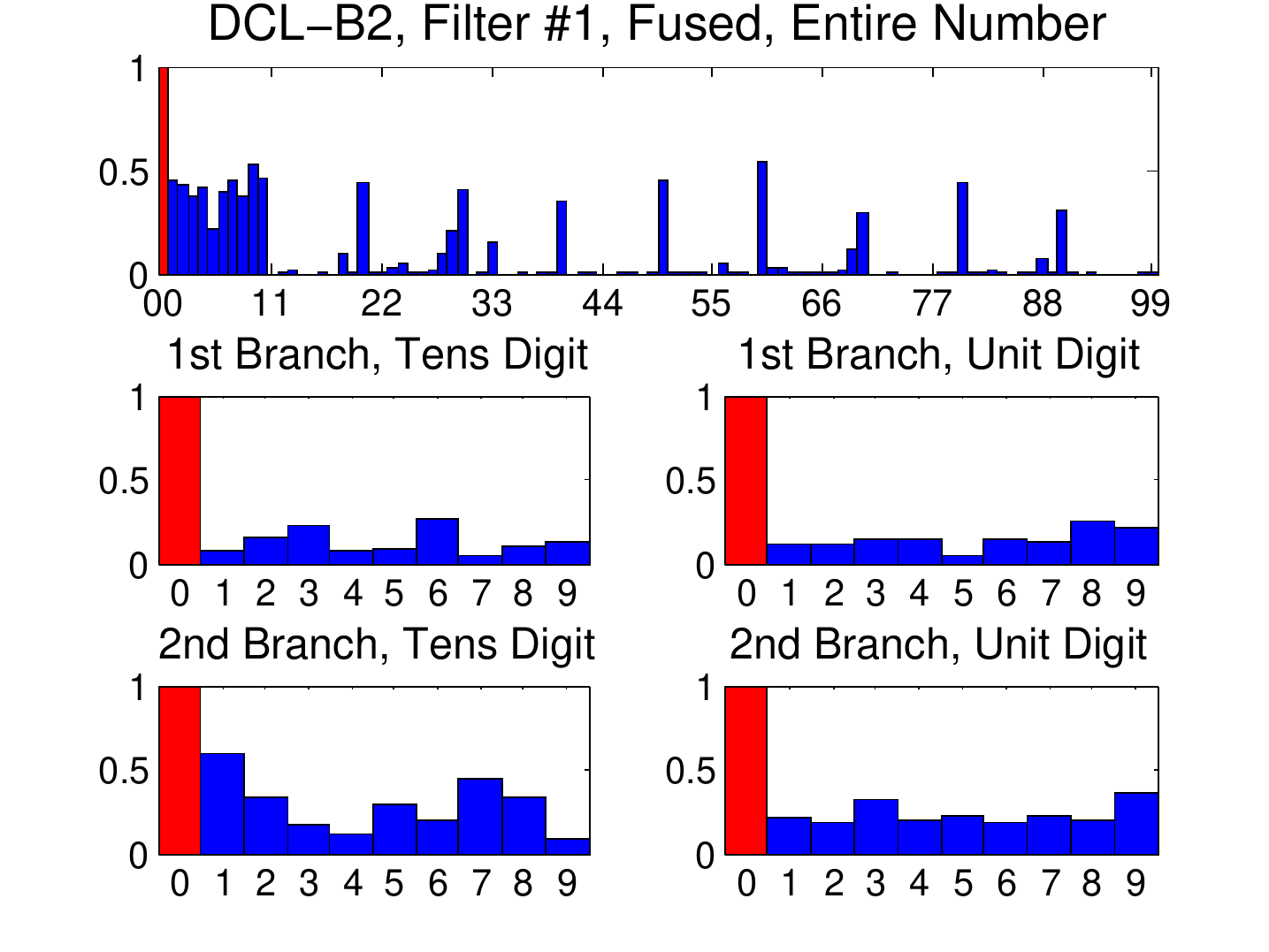}
\end{center}
\caption{
    Average neural responses over $100$ classes using two filters (best viewed on color PDF).
    Each fused filter gets high responses in a specific class (marked in red).
    In the subplot of each branch (separate filter), the red bar indicates the highest response,
    and the green bar, if necessary, indicates a different unit that contributes to the highest response in fusion.
}
\label{Fig:ComplementaryKnowledge}
\end{figure*}

We show that DCL is able to learn complementary visual knowledge.
To this end, the {\bf DCL-A2} and {\bf DCL-B2} models trained on the {\bf MNIST-II-05} dataset are investigated.
We show some statistics of the first filter in each DCL module,
including the fused response $z$ and the individual responses $v^{\left(1\right)}$ and $v^{\left(2\right)}$.
We feed all $10\rm{,}000$ testing images to these filters,
obtain $10\rm{,}000$ results for $z$, $v^{\left(1\right)}$ and $v^{\left(2\right)}$, respectively.
These values grouped using the ground-truth label (the number, the tens digit and the unit digit).
In Figure~\ref{Fig:ComplementaryKnowledge}, we plot the average responses of $z$ on different numbers ($100$ classes)
and the average responses of $v^{\left(1\right)}$ and $v^{\left(2\right)}$ on different digits ($10$ classes).

We can observe that each fused filter strongly responds to a specific class,
and these strong responses come from the individual branches.
For example, the fused filter in {\bf DCL-A2} learns the visual category 68,
thus the corresponding individual filters can learn 6 in the tens digit and 8 in the unit digit, respectively.
This makes is possible to implicitly decompose the learning task into smaller and easier ones.

\subsection{Generic Image Classification}
\label{Experiments:Generic}

We show that Deep Collaborative Learning also works well in generic image classification tasks.
We evaluate it on four popular datasets,
namely {\bf SVHN}, {\bf CIFAR10}, {\bf CIFAR100}, and the large-scale {\bf ILSVRC2012} database.

\subsubsection{The SVHN and CIFAR Datasets}
\label{Experiments:Generic:SmallDatasets}

{\bf SVHN} (Street View House Numbers)~\cite{Netzer_2011_Reading} is a large collection of $32\times32$ RGB images,
{\em i.e.}, $73\rm{,}257$ training samples, $26\rm{,}032$ testing samples, and $531\rm{,}131$ extra training samples.
We preprocess the data as in the previous work~\cite{Netzer_2011_Reading},
{\em i.e.}, selecting $400$ samples per category from the training set as well as $200$ samples per category from the extra set,
using these $6\rm{,}000$ images for validation, and the remaining $598\rm{,}388$ images as training samples.
We also use Local Contrast Normalization (LCN) for data preprocessing~\cite{Goodfellow_2013_Maxout}.

{\bf CIFAR10} and {\bf CIFAR100}~\cite{Krizhevsky_2009_Learning}
are both subsets drawn from the $80$-million tiny image database~\cite{Torralba_2008_80}.
There are $50\rm{,}000$ images for training, and $10\rm{,}000$ images for testing, all of them are $32\times32$ RGB images.
{\bf CIFAR10} contains $10$ basic categories, and {\bf CIFAR100} divides each of them into a finer level.
In both datasets, training and testing images are uniformly distributed over all the categories.
We use exactly the same network configuration as in the {\bf SVHN} experiments,
and add left-right image flipping into data augmentation with the probability $50\%$.

We use three network structures.
The first one is a variant of the {\bf LeNet} model used in {\bf MNIST} experiments.
The network structure contains three convolutional layers, three pooling layers and two fully-connected layers:
\begin{spverbatim}
C5(P2)@32-MP3(S2)-C5(P2)@64-MP3(S2)-
C5(P2)@128-MP3(S2)-FC512-D0.5-OUT.
\end{spverbatim}
\noindent
We apply $120$ training epochs with learning rate $10^{-3}$, followed by $20$ epochs with learning rate $10^{-4}$,
and another $10$ epochs with learning rate $10^{-5}$.

The second one is named the {\bf BigNet}, which is borrowed from~\cite{Nagadomi_2014_Kaggle}.
In {\bf CIFAR} datasets, we randomly flip the image with $50\%$ probability.
We train the {\bf BigNet} using $6\times10^6$ samples with learning rate $10^{-2}$,
followed by $3\times10^6$ samples with learning rate $10^{-3}$ and $1\times10^6$ samples with learning rate $10^{-4}$, respectively.
We report a $7.88\%$ error rate on {\bf CIFAR10}, comparable to the original version~\cite{Nagadomi_2014_Kaggle},
which uses a very complicated way of data preparation and augmentation to get a $6.68\%$ error rate.
Training the original version~\cite{Nagadomi_2014_Kaggle} requires $6$ hours, while our model needs only $1$ hour.
The final baseline, Wide Residual Net ({\bf WRN})~\cite{Zagoruyko_2016_Wide},
takes the advantage of deep residual learning~\cite{He_2016_Deep},
and uses a larger number of convolutional kernels and a smaller number of layers.
We follow the original implementation to train the $16$-layer WRN, which takes around $6$ hours to complete a single model.

\newcommand{\colwidth}{1.0cm}
\begin{table}[t]
\begin{center}
\begin{tabular}{|l||R{\colwidth}|R{\colwidth}|R{\colwidth}|}
\hline
{}                                                   & {\bf SVHN}       & {\bf CF10}       & {\bf CF100}      \\
\hline\hline
Zeiler {\em et.al}~\cite{Zeiler_2013_Stochastic}     & $ 2.80$          & $15.13$          & $42.51$          \\
\hline
Goodfellow {\em et.al}~\cite{Goodfellow_2013_Maxout} & $ 2.47$          & $ 9.38$          & $38.57$          \\
\hline
Lin {\em et.al}~\cite{Lin_2014_Network}              & $ 2.35$          & $ 8.81$          & $35.68$          \\
\hline
Lee {\em et.al}~\cite{Lee_2015_Deeply}               & $ 1.92$          & $ 7.97$          & $34.57$          \\
\hline
Liang {\em et.al}~\cite{Liang_2015_Recurrent}        & $ 1.77$          & $ 7.09$          & $31.75$          \\
\hline
Lee {\em et.al}~\cite{Lee_2016_Generalizing}         & $ 1.69$          & $ 6.05$          & $32.37$          \\
\hline
Xie {\em et.al}~\cite{Xie_2016_Geometric}            & $\mathbf{ 1.67}$ & $ 5.31$          & $25.01$          \\
\hline
Huang {\em et.al}~\cite{Huang_2016_Deep}             & $ 1.75$          & $\mathbf{ 5.25}$ & $\mathbf{24.98}$ \\
\hline\hline
{\bf LeNet} (w/o DCL)                                & $ 4.21$          & $14.18$          & $44.77$          \\
\hline
{\bf LeNet} (w/ DCL)                                 & $\mathbf{ 3.80}$ & $\mathbf{13.55}$ & $\mathbf{42.81}$ \\
\hline
{\bf BigNet} (w/o DCL)                               & $ 2.19$          & $ 7.88$          & $31.03$          \\
\hline
{\bf BigNet} (w/ DCL)                                & $\mathbf{ 2.03}$ & $\mathbf{ 7.46}$ & $\mathbf{29.83}$ \\
\hline
{\bf WRN} (w/o DCL)                                  & $ 1.77$          & $ 5.54$          & $25.52$          \\
\hline
{\bf WRN} (w/ DCL)                                   & $\mathbf{ 1.69}$ & $\mathbf{ 5.37}$ & $\mathbf{25.10}$ \\
\hline
\end{tabular}
\caption{
    Comparison of the recognition error rates ($\%$) with the state-of-the-arts.
    We apply data augmentation on all these datasets.
}
\label{Tab:SmallDatasets}
\end{center}
\end{table}

Results are summarized in Table~\ref{Tab:SmallDatasets}.
We add DCL (two branches, each of which has $1/4$ of the original number of filters)
to replace the first fully-connected layer of all three baselines.
Consistent accuracy gain is observed.
With {\bf LeNet}, the relative error rate drops are $9.74\%$, $4.44\%$ and $4.38\%$ on the three datasets;
with {\bf BigNet}, these numbers are $7.31\%$, $5.33\%$ and $3.87\%$;
with {\bf WRN}, these numbers are $4.52\%$, $3.07\%$ and $1.65\%$.
These experiments verify that DCL generalizes well to both shallow nets ($5$-layer {\bf LeNet})
and deep nets ($11$-layer {\bf BigNet} and $16$-layer {\bf WRN}).
Note that DCL achieves accuracy gain with fewer network parameters,
{\em e.g.}, with $10$ output nodes,
the number of trainable weights in {\bf BigNet} is shrunk from $5.74\mathrm{M}$ to $5.09\mathrm{M}$ ($\mathbf{11.41\%}$ {\bf fewer}).

\subsubsection{The ILSVRC2012 Dataset}
\label{Experiments:Generic:ILSVRC}

Finally, we evaluate our model on the {\bf ImageNet} large-scale visual recognition task
(the {\bf ILSVRC2012} dataset~\cite{Russakovsky_2015_ImageNet} with $1000$ categories).
We use the {\bf AlexNet} provided by the {\bf CAFFE} library~\cite{Jia_2014_CAFFE}, which is abbreviated as:
\begin{spverbatim}
C11(S4)@96-MP3(S2)-C5(S1P2)@256-MP3(S2)-
C3(S1P1)@384-C3(S1P1)@384-C3(S1P1)@256-
MP3(S2)-FC4096-D0.5-FC4096-D0.5-FC1000.
\end{spverbatim}
\noindent
The input image is of size $227\times227$, randomly cropped from the original $256\times256$ image.
Following the setting of {\bf CAFFE}, a total of $450\rm{,}000$ mini-batches (approximately $90$ epochs) are used for training,
each of which has $256$ image samples, with the initial learning rate $10^{-2}$, momentum $0.9$ and weight decay $5\times10^{-4}$.
The learning rate is decreased to $1/10$ after every $100\rm{,}000$ mini-batches.

We replace the original {\em fc-6} layer ($4096$ filters) which two DCL branches, each of which has $1024$ filters.
With DCL, the top-$1$ and top-$5$ recognition error rates are $42.98\%$ and $19.69\%$, respectively.
Comparing to the original rates ($43.19\%$ and $19.87\%$), DCL relatively decreases them by about $0.5\%$ and $0.9\%$, respectively.
We emphasize that the accuracy gain is not as small as it seems,
especially when the number of parameters decreases from $62.35\mathrm{M}$ to $51.86\mathrm{M}$ ($\mathbf{16.82\%}$ {\bf fewer})
and the average training time per $20$ iterations decreases from $6.04\mathrm{s}$ to $5.67\mathrm{s}$ ($\mathbf{6.13\%}$ {\bf less}).

Although our algorithm is only tested on {\bf AlexNet}, we believe it can be applied to other models,
such as {\bf VGGNet}~\cite{Simonyan_2015_Very}, {\bf GoogleNet}~\cite{Szegedy_2015_Going} and Deep Residual Nets~\cite{He_2016_Deep}.

\section{Conclusions}
\label{Conclusions}

This paper presents Deep Collaborative Learning (DCL), a generalized module which can be plugged into a large family of networks.
A DCL module consists of two stages,
in which we first build some intermediate branches and then fuse them at each spatial position to consider feature co-occurrence.
DCL allows us to construct an exponentially large visual vocabulary with linear complexity,
which, in practice, reduces the number of trainable parameters of each model, and alleviates the risk of over-fitting.
In experiments, DCL significantly outperforms the baseline model on a series of multi-digit number datasets,
and generalizes well to a wide range of generic image classification tasks.
We also verify that DCL is able to learn complementary information in different branches.

We learn from DCL that a large filter set can be simulated by several small filter banks.
In the current state, DCL works in a fixed decomposition-fusion manner.
It would be very interesting to allow neural connections between some small convolutional layers.
Meanwhile, other visual tasks, including detection, segmentation, {\em etc.}, may also benefit from the DCL.
The exploration of these topics is left for future work.

{\small
\bibliographystyle{ieee}
\bibliography{egbib}
}

\end{document}